\documentclass[runningheads]{llncs}

\usepackage{eccv}

\usepackage{eccvabbrv}

\usepackage{graphicx}
\usepackage{multirow}
\usepackage{arydshln}
\usepackage{array}
\usepackage{gensymb}
\usepackage{subcaption}
\usepackage[export]{adjustbox}
\usepackage{booktabs}

\usepackage{colortbl}
\usepackage[accsupp]{axessibility}  

\usepackage[pagebackref,breaklinks,colorlinks,citecolor=eccvblue]{hyperref}

\usepackage{orcidlink}

\begin{document}

\title{MPSI: Mamba enhancement model for pixel-wise sequential interaction Image Super-Resolution} 

\titlerunning{Abbreviated paper title}

\author{Yuchun He\inst{1}\orcidlink{0009-0003-1457-8719} \and
Yuhan He\inst{1}\orcidlink{0009-0003-2725-9178}}

\authorrunning{Y. He et al.}

\institute{University of Sydney, Camperdown NSW 2050, Australia
\email{campus.assist@sydney.edu.au}\\
\url{https://www.sydney.edu.au/}}

\maketitle

\begin{abstract}
    Single image super-resolution (SR) has long posed a challenge in the field of computer vision. While the advent of deep learning has led to the emergence of numerous methods aimed at tackling this persistent issue, the current methodologies still encounter challenges in modeling long sequence information, leading to limitations in effectively capturing the global pixel interactions. To tackle this challenge and achieve superior SR outcomes, we propose the Mamba pixel-wise sequential interaction network (MPSI), aimed at enhancing the establishment of long-range connections of information, particularly focusing on pixel-wise sequential interaction. We propose the Channel-Mamba Block (CMB) to capture comprehensive pixel interaction information by effectively modeling long sequence information. Moreover, in the existing SR methodologies, there persists the issue of the neglect of features extracted by preceding layers, leading to the loss of valuable feature information. While certain existing models strive to preserve these features, they frequently encounter difficulty in establishing connections across all layers. To overcome this limitation, MPSI introduces the Mamba channel recursion module (MCRM), which maximizes the retention of valuable feature information from early layers, thereby facilitating the acquisition of pixel sequence interaction information from multiple-level layers. Through extensive experimentation, we demonstrate that MPSI outperforms existing super-resolution methods in terms of image reconstruction results, attaining state-of-the-art performance.
\end{abstract}

\section{Introduction}
\label{sec:intro}
Single image super-resolution (SR) \cite{SR} stands a low-level task within computer vision, aspiring to recuperate intricate high-frequency details from low-resolution (LR) images to reconstruct high-resolution (HR) counterparts. In practical applications, SR holds significant relevance, particularly in domains such as image restoration and video enhancement \cite{cat, videoSR}. In recent years, advancements in SR techniques leveraging deep learning have primarily bifurcated into two avenues: methodologies based on convolutional neural networks (CNN) \cite{SISR, zhang2018image, zhang2021plugandplay}, and Transformer-based approaches  \cite{swinir, ESRT, ELAN}. While SR models based on CNN \cite{CNN} can enhance image details, the inherent limitation of CNN lies in establishing pixel feature dependencies, thus constraining the capacity for detailed restoration. The Transformer \cite{transformer} employs a self-attention mechanism to facilitate dependency modeling and feature connectivity, demonstrating superior performance over conventional CNN-based structures in image SR tasks. However, Transformer-based SR models are constrained by windows, thereby impeding their ability to model long-range dependencies within long sequence information. ELAN \cite{ELAN} endeavors to address the problem of modeling long-range dependencies in features by employing long-range attention. However, the dependencies captured by ELAN remain confined within the window, thus presenting limitations in the analysis of broader long-range contexts. Although DAT \cite{DAT} uses channel attention to exploit global information, the Transformer's constraints in modeling long sequence information limit the understanding of global context.

We introduce the Mamba pixel-wise sequential interaction network (MPSI) for the SR task to augment the sequential relation of pixel information. Then we propose Channel-Mamba Block (CMB) which integrates the Mamba \cite{mamba} structure and leverages its state space model with a selection mechanism to model long sequence information, thus augmenting its comprehension of global pixel information. CMB utilizes the Dual Direction Bi-Mamba module (DDBM), which addresses the limitation of one-way modeling with Mamba, to conduct a bidirectional analysis of feature sequences and establish comprehensive global pixel sequence interaction relationships. MPSI proposes the Spatial Transformer Block (STB) to extract the primary spatial feature sequence, then deploy the CMB to extract the channel-wise information, efficiently aggregating the image details captured in spatial and channel dimensions, enriching the spatial expression of each feature map and long-range dependence.

In deep learning networks, each layer processes features uniquely due to variations in weights and biases. However, as the network deepens, the influence of earlier layers on the outcomes diminishes \cite{resnet}. Transformer-based networks such as SwinIR establish connections between layers so that the front layers can obtain more effective optimization and affect the final result. However, those approaches affect the symmetry of the network and may lead to over-fitting \cite{denseNEt}. To mitigate this issue, we propose the Mamba channel recursion module (MCRM), drawing inspiration from the principles of SENet \cite{SENet}. MCRM recursively analyzes the features of each network layer and exerts greater influence on the output features in the form of channel weights through the gate structure, further enhancing the pixel-wise sequential interaction. As a result, the model optimally exploits information obtained from deep feature extraction, enhancing overall feature representation. Our primary contributions are summarized as follows:
\begin{enumerate}
    \item We propose Channel-Mamba Block (CMB), a Mamba-based enhancement framework that effectively models the long-range dependence of the long sequence information to enhance the pixel-wise sequential interaction.
    \item We propose the Mamba channel recursion module (MCRM), which holistically analyzes the features of each layer and summarizes the channel weights. This can compensate for the influence of early layers on the pixel sequence interactions, and enhance the model’s ability to analyze features.
    \item Our proposed MPSI notably surpasses the existing methods, attaining state-of-the-art performance.
\end{enumerate}

\section{Related Work}

\subsection{Super Resolution}
Super-resolution constitutes a pivotal facet within the realm of image restoration, where deep learning has demonstrated remarkable efficacy. Leveraging deep neural networks, particularly convolution-based architectures, has yielded significant advancements in super-resolution tasks. Notably, the Super-Resolution Convolutional Neural Network (SRCNN) \cite{SISR} epitomizes this progress, comprising three integral components: Patch Extraction and Representation, Non-Linear Mapping, and Reconstruction. Each segment harnesses the power of 2D convolutions and ReLU activation functions for feature extraction and enhancement. Building upon the foundation laid by SRCNN, the Coarse-to-Fine Super-Resolution CNN (CFSRCNN) \cite{CFSRCNN} emerged, incorporating diverse modules to capture complementary contextual information and bolster super-resolution outcomes. Furthermore, the integration of a ResNet structure \cite{resnet} serves to mitigate gradient vanishing and explosion issues, thus enhancing training stability. Efforts are underway to optimize super-resolution networks by reducing parameter counts without compromising performance. Prominent examples include the PAN model proposed by Hengyuan et al. \cite{PAN} and the Lightweight Enhanced SR CNN (LESRCNN) \cite{LESRCNN} introduced by Chunwei et al. These endeavors aim to bolster training efficiency and computational efficacy. The advent of Transformer architecture has ushered in a paradigm shift in super-resolution research, expanding beyond convolution-centric approaches. Self-attention mechanisms present in Transformers offer new avenues for exploration and innovation in super-resolution tasks, promising further breakthroughs in this dynamic field.

\begin{figure}[t]
    \centering
    \includegraphics[width=1\textwidth]{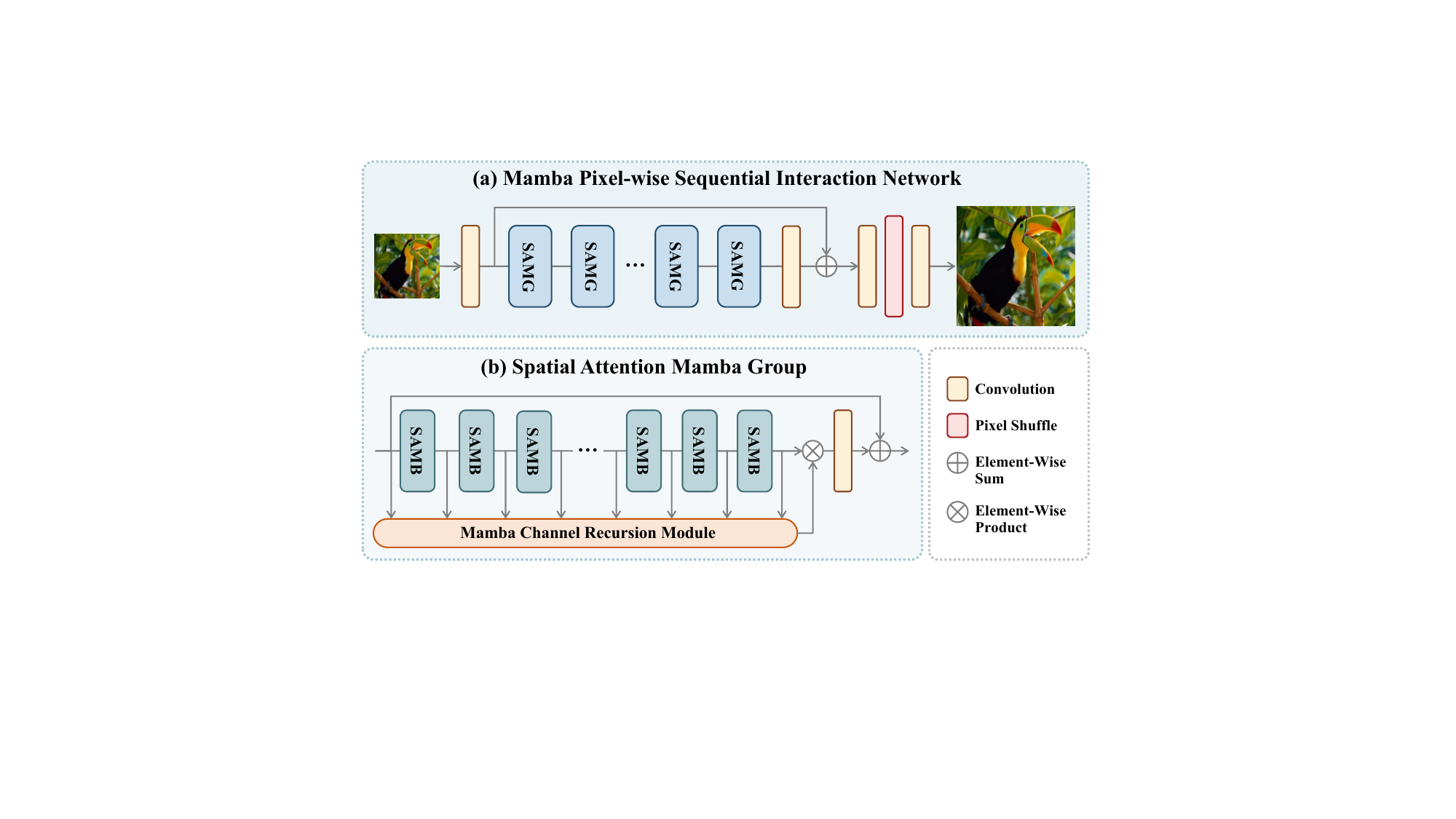}
    \caption{Architecture of MPSI net work and SAMG.}
    \label{MPSI}
\end{figure}

\subsection{Transformer}
The self-attention mechanism in Transformer \cite{transformer} brings more possibilities to deep learning. The remarkable performance of Transformers in natural language processing \cite{GPT, ctrl, BERT, label2label} has spurred considerable interest among researchers regarding their applicability in visual tasks \cite{zou2024contourlet, li2025contourlet, li2023text}. For instance, Vision Transformer (ViT) \cite{Vit} leverages Transformer architecture to address image classification tasks. Swin Transformer \cite{swin} introduces shifted windows, enabling Transformers to perform self-attention calculations on images with interactive windows, thereby enhancing adaptability to diverse visual tasks. SwinIR \cite{swinir} draws inspiration from Swin Transformer and constructs a block based on its architecture to conduct cross-window local attention calculations on images. By amalgamating the strengths of Transformer and CNN, SwinIR effectively tackles single-image super-resolution tasks, yielding commendable results. DaViT \cite{DaViT} introduced a Transformer architecture amalgamating spatial window self-attention and channel group self-attention to facilitate comprehensive information capture for addressing visual tasks. Building upon DaViT's framework, Dual Aggregation Transformer (DAT) \cite{DAT} devised Dual Spatial Transformer and Dual Channel Transformer blocks. Additionally, DAT incorporated adaptive interaction modules and a spatial-gate feed-forward network structure to construct a single-image super-resolution model. 

Window self-attention-based models frequently necessitate architectures with high-channel and high-depth configurations to attain superior performance. Conversely, when employing relatively lightweight models characterized by low-channel and low-depth designs, the resulting generation quality often tends to be suboptimal. Therefore, we propose leveraging the Mamba structure to construct a lightweight model, aiming to enhance the effectiveness of super-resolution (SR) models under low-channel and low-depth conditions.

\subsection{Mamba}
Mamba \cite{mamba} is a novel feature sequence processing module inspired by the structured state-space sequence model \cite{SSMs_1, SSMs_2}. Unlike self-attention, Mamba's learnable variables do not increase with the length of the feature sequence but are related to the size of a single feature in the sequence. It integrates a selection mechanism to handle discrete, information-dense data. Models related to Mamba demonstrate proficiency in analyzing remote information dependencies. One advantage is that the length of the input feature sequence is not limited, and the feature sequence does not require position embedding. Mamba has shown great effectiveness and has very good performance in natural language processing.

In addition to natural language processing, Mamba has begun to find applications in various fields, including graph processing \cite{gmamba} and image processing \cite{panmamba}. Unlike attention mechanisms, which are sensitive to the length of the feature sequence, Mamba focuses more on the order of feature sequences. In the domain of pan-sharpening, Xuanhua et al. attempted to apply Mamba and proposed the Pan-mamba network model \cite{panmamba}. At the same time, Wang et al. proposed Mamba-unet use Mamba's long sequence and global information understanding capabilities to improve the effect of medical image segmentation and achieved better results than Unet \cite{Unet} and Swin-Unet \cite{Swin-unet}. This study proposes the integration of the Mamba structure into the architecture of an SR model to enhance the model's capability to capture long-range feature dependencies. Our objective is to propose the MPSI to comprehend long sequences and global information, enabling the model to capture dependencies among long-range features in images. This aims to enhance the sequential relation of pixel information.

\begin{figure}[t]
    \centering
    \includegraphics[width=1\textwidth]{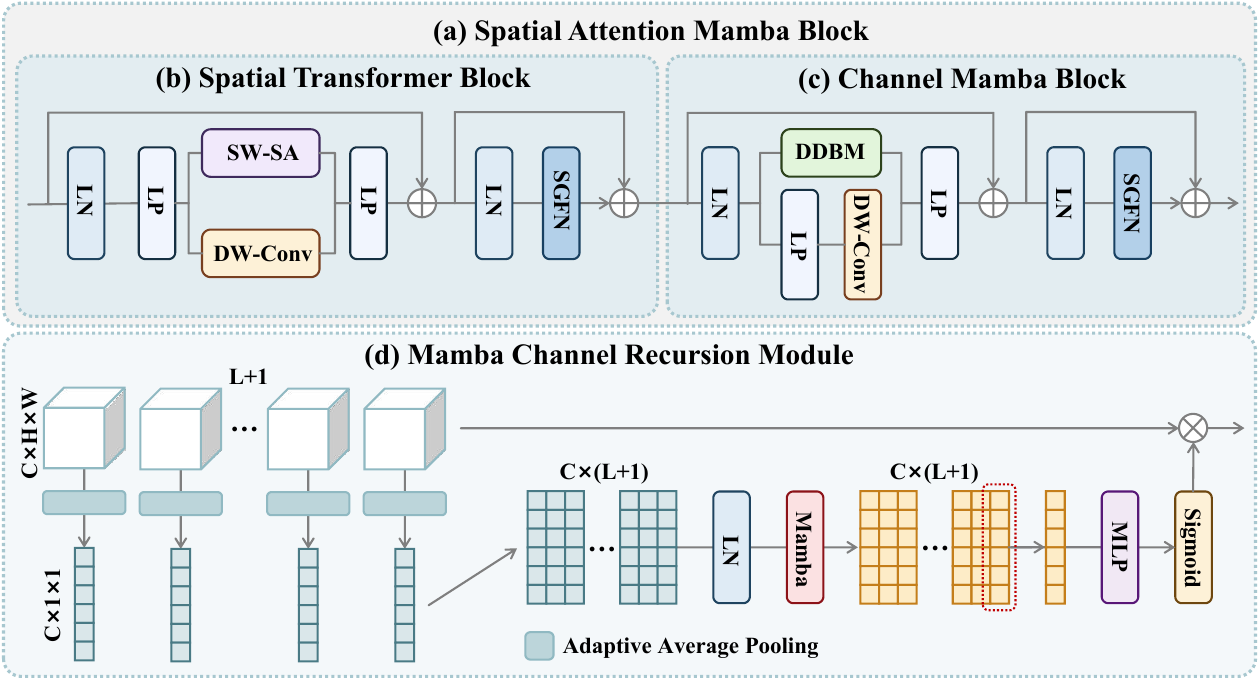}
    \caption{Components of SAMG. The STB and CMB are the components of SAMB. Figures b, c, and d show the structures of STB, CMB, and MCRM respectively.}
    \label{SAMG}
\end{figure}

\section{Method}
\subsection{Architecture}
The Mamba pixel-wise sequential interaction network (MPSI) comprises three main components: shallow feature extraction, deep feature extraction, and image reconstruction. The structure of MPSI is shown in Figure \ref{MPSI}. The input to the network is a batch of LR images $\mathit{I_{LR}}\in\mathbb{R}^{B\times 3\times H\times W}$, where $B$ denotes the batch size, $\mathit{H}$ represents the image height, and $\mathit{W}$ signifies the image width. The convolutional layer within the shallow feature extraction generates features $\mathit{F_S}\in\mathbb{R}^{B\times C\times H\times W}$, with $\mathit{C}$ denoting the number of channels. 

In the subsequent stage, the shallow feature is input into the deep feature extraction module. The feature's shape is transformed into $S\in \mathbb{R}^{B\times HW \times C}$ before undergoing deep feature extraction. This configuration facilitates the treatment of features as sequences, facilitating subsequent self-attention and Mamba operations. The deep feature extraction module comprises a series of spatial attention Mamba groups (SAMG), the number of which is denoted by $\mathit{N}$. Each SAMG comprises several spatial attention Mamba blocks (SAMB), with each SAMB consisting of a pair of Spatial Transformer Blocks (STB) and Channel-Mamba Blocks (CMB). The quantity of SAMBs in a SAMG is expressed as $\mathit{M}$, resulting in a total depth of MPSI expressed as $2NM$. The Mamba channel recursion module (MCRM) operates on features generated by SAMBs within a SAMG, enhancing feature distinctions at the pixel level. The resulting features after deep feature extraction are denoted as $\mathit{F_D}\in\mathbb{R}^{B\times C\times H\times W}$.

Finally, the deep features undergo image reconstruction to generate an HR image. Through pixel shuffling, the deep features are upsampled to produce the resulting image $\mathit{I_{HR}}\in\mathbb{R}^{B\times 3\times H_n\times W_n}$, where $H_n$ and $W_n$ denote the height and width of the resulting image, respectively. Our model is optimized by MSE loss and perceptual loss.
\begin{figure}[t]
    \centering
    \includegraphics[width=1\textwidth]{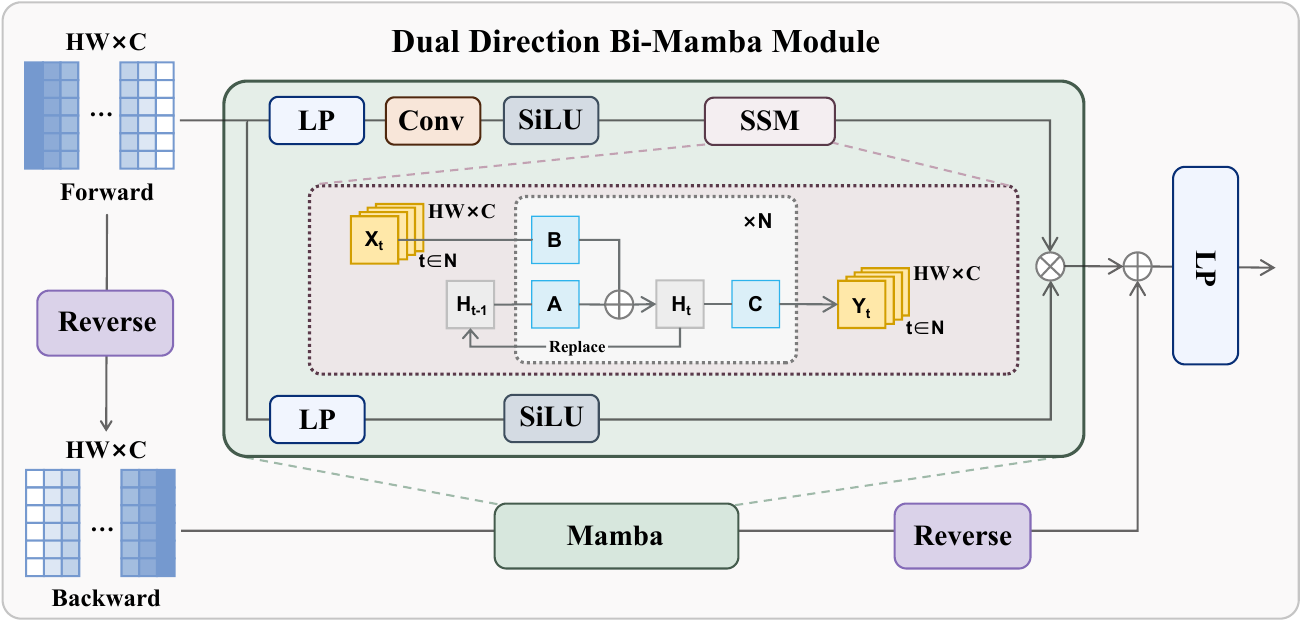}
    \caption{Architecture of DDBM. The initial value of $H_{t-1}$ is a zero matrix. N is the number of features in the feature sequence.}
    \label{mamba}
\end{figure}
\subsection{Spatial Attention Mamba Group}
The spatial attention Mamba group (SAMG) serves as the cornerstone for deep feature extraction, embodying a pivotal architecture depicted in Figure \ref{MPSI} (b), with detailed constituents outlined in Figure \ref{SAMG}. LP in the figure refers to linear projection, which can be expressed as $LP(X) = WX + b$. Each SAMG comprises multiple spatial attention Mamba blocks (SAMB) and a single Mamba channel recursion module (MCRM). Within the SAMB, the Spatial Transformer Block (STB) and the Channel-Mamba Block (CMB) synergistically collaborate, drawing inspiration from the efficacy of Dual Aggregation Transformer (DAT) networks \cite{DAT}, to enable comprehensive feature analysis across spatial and channel dimensions. The input and output dimensions of each SAMB remain invariant, spanning $\mathbb{R}^{B\times HW\times C}$. Both STB and CMB harness the spatial-gate feed-forward network (SGFN) \cite{DAT} to facilitate meticulous feature aggregation and refinement, ensuring robust performance across diverse spatial and channel contexts. This architecture epitomizes the fusion of cutting-edge feature sequence analysis mechanisms and modular design principles, culminating in an effective framework for deep feature extraction in visual processing tasks.

\subsubsection{Spatial Transformer Block}
The Spatial Transformer Block (STB) block's main function is to analyze features' spatial relationships. As shown in Figure \ref{SAMG} (b), Its main mechanism is spatial window self-attention(SW-SA) \cite{DAT, DaViT}. The way SW-SA handles features in SAT is very similar to DAT \cite{DAT}. Assume the input feature is $S$. After layer normalization, STB first obtains $Q$, $K$, and $V$ through linear projection with bias.
SW-SA in STB uses windows to organize $Q$, $K$, and $V$ into sequences. The size of the window is $M\times N$. The shape of $Q$, $K$, and $V$ become $\mathbb{R}^{B\times \frac{HW}{MN}\times MN \times C}$. Afterward, $Q$, $K$, and $V$ are divided into multiple heads, the number of heads is $H$. Then the shape of each element is $H\times \frac{C}{H}$.
While computing SW-SA, depth-wise convolution (dw-conv) is employed on V. Subsequently, STB amalgamates features from SW-SA and dw-conv, followed by a linear projection with weights denoted by $W$. Upon addition of $S$, STB obtains a temporary feature $\mathit{S_{tmp}}\in\mathbb{R}^{B\times HW\times C}$. The process is shown as:
\begin{equation}
    S_{tmp}=LP(SWSA(Q,K,V) + DWcov(V)) + S.
    \label{eq:stb1}
\end{equation}
Finally, $\mathit{S_{tmp}}$ undergoes layer normalization and is passed through an SGFN block. Following the addition of $\mathit{S_{tmp}}$, STB obtains the output feature $\mathit{S_{stb}}$. The calculation process is as:
\begin{equation}
    S_{stb}=SGFN(LN(S_{tmp})) + S_{tmp}
    \label{eq:stb2}
\end{equation}

\subsubsection{Channel-Mamba Block}
The primary function of the Channel-Mamba Block (CMB) is to conduct feature analysis at the channel level. The structure of CMB is shown in Figure \ref{SAMG} (c). To achieve this, the CMB introduces the Dual Direction Bi-Mamba module (DDBM), as illustrated in Figure \ref{mamba}. Because image features should not have a fixed processing order, DDBM processes forward and backward feature sequences simultaneously to solve the problem of Mamba being sensitive to the order of feature sequences.

For the entire CMB, the input feature $S$ undergoes layer normalization before it enters into two parts simultaneously, one is the DDBM module and the other is the linear projection and DW-conv module \cite{DAT}. The results are then element-wise summed and added to the initial feature $S$ yielding $\mathit{S_{tmp}}$. The process is shown as:
\begin{equation}
    S_{tmp}=LP(DDBM(S) + DWcov(LP(S))) + S.
    \label{eq:cmb}
\end{equation} 
Subsequently, $\mathit{S_{tmp}}$ undergoes another layer of normalization and the SGNF block. Upon summation of $\mathit{S_{tmp}}$, the resulting feature is denoted as $\mathit{S_{cmb}}$. This process is the same with STB.

In DDBM, the input feature sequence $S = \{x_1,\ x_2,\ ...,\ x_k\}$ is copied and reverse-sorted to obtain $S_r$, where $k=H_nW_n$. Subsequently, $S$ and $S_r$ enter two separate Mamba blocks. The Space State Model (SSM) \cite{mamba} can use matrix groups $(A, B, C)$ and potential states $h$ to connect features in the feature sequence to achieve long-range connections between features, thereby achieving better super-resolution results. The functionality of SSM is illustrated as:\\ 
\begin{minipage}{0.5\textwidth}
    \centering
    \begin{subequations}\label{a}
    \begin{align}
        h_t &= Ah_{t-1}+Bx_t. \label{ssm1} \\
        y_t &= Ch_t. \label{ssm2}
    \end{align}
    \end{subequations}
\end{minipage}%
\begin{minipage}{0.5\textwidth}
    \centering
    \begin{subequations}\label{b}
    \begin{align}
        K &= \{CB,\ CAB,\ ...,\ CA^kB\}. \label{ssm3} \\
        Y &= KS. \label{ssm4}
    \end{align}
    \end{subequations}
\end{minipage}\\\\
In addition to the SSM block, Mamba also includes its linear projection, convolution and other structures, as depicted in Figure \ref{mamba}. Subsequent to the Mamba block, CMB produces two feature sequences $Y_f$ and $Y_b$. Reversing the arrangement of $Y_b$ and performing element-wise addition with the two sets of outputs yields $Y_s$. After undergoing a linear projection, DDBM converts $Y_s$ to $S_{out}$ as the output. The calculation process of DDBM is illustrated as:
\begin{equation}
\begin{gathered}
Mamba(S) = SiLU(LP(S))\otimes SSM(SiLU(Conv(LP(S))), \\
Y_f = Mamba(S),\ Y_b = revers(Mamba(S_r)), \\
S_{out} = LP(Y_f \oplus Y_b).
\end{gathered}
\label{DDBM}
\end{equation}

\subsection{Mamba channel recursion module}

The structure of the Mamba channel recursion module (MCRM) is depicted in Figure \ref{SAMG} (d), which ignores the batch size. For this process to work efficiently, the shape of the feature is temporarily changed from $S\in \mathbb{R}^{B\times HW \times C}$ to $F\in\mathbb{R}^{B\times C\times H\times W}$. Assuming $L\in \mathbb{R}$ denotes the number of spatial attention Mamba blocks (SAMB), MCRM conducts adaptive average pooling on features $F_{layers} = \{F_1,\ F_2,\ ...,\ F_L,\ F_{L+1}\}$ after input and each SAMB, along with a feature preceding the spatial attention Mamba group (SAMG). Subsequently, SAMG acquires $L+1$ features $P = \{X_1^p,\ X_2^p,\ ...,\ X_L^p,\ X_{L+1}^p\}$, each with dimensions $\mathbb{R}^{B\times C\times 1}$. This process can be represented as below, where $X_t^p$ denotes each element within $P$.
\begin{equation}
    X_t^p = AdaptiveAvgPool(F_t).
    \label{pool}
\end{equation}
After layer normalization of $P$, it proceeds into the Mamba module which is the key for recursive analysis, resulting in $P_{out}\in \mathbb{R}^{B\times L\times C}$. The last column $X_{last}\in \mathbb{R}^{B\times C}$ of $P_{out}$ then undergoes processing through a multilayer perceptron (MLP) and sigmoid activation function to obtain layer weight $X_w\in \mathbb{R}^{B\times C}$. This weighted feature, $X_w$, is then multiplied with the last feature $F_{L+1}$ of $F_{layers}$, resulting in $F_{out}$ as the output of SAMG. The process is illustrated as:
\begin{equation}
    \begin{aligned}
        & X_w = Sigmoid(MLP(X_{last})), \\
        & F_{out} = F_{L+1} \otimes X_w.
    \end{aligned}
    \label{MCRM}
\end{equation}



\section{Experiments}
In this section, we conduct quality assessment experiments on MPSI and compare it with previous SR models. Additionally, ablation experiments were conducted to systematically assess the contribution of each component within MPSI.
\subsection{Experimental setup}
\subsubsection{Dataset and evaluation metrics.} We combined two distinct datasets for training purposes, namely DIV2K \cite{DIV2K} and Flickr2K \cite{Flickr2K}. We acquired LR images through degradation. For validation, we utilized a comprehensive set comprising Set5 \cite{Set5}, Set14 \cite{Set14}, B100 \cite{B100}, Urban100 \cite{Urban100}, and Manga109 \cite{Manga109}. HR images were degraded into LR images using bicubic degradation, with upscaling factors of $\times$2, $\times$3, and $\times$4. Our evaluation of the super-resolution outcomes was conducted using two widely recognized metrics, namely PSNR and SSIM. We conduct comparative analyses with several other SR models to evaluate the performance capabilities of the proposed MPSI.
\subsubsection{Implementation Details.} For MPSI, the deep feature extraction component consists of 1 SAMG containing 9 SAMBs, and each SAMB has 1 STB and 1 CMB. Each attention head is configured with 6 heads, the channel dimension of the SGFN is set to 60, and the expansion factor is set to 2. The window dimensions are specified as 8x32. For DDBM in the CMB, the settings of two Mamba blocks are the same, where the SSM state expansion factor is set to 32, the local convolution width is 3, and the block expansion factor is 4. In the MCRM, the SSM state expansion factor is set to 64, the local convolution width set to 4, and the block expansion factor set to 2.
\subsubsection{Training details.} Our model undergoes training with a batch size of 8 and a patch size of 64x64. The training process comprises a total of 500K iterations, employing the Adam optimizer to minimize the $L_1$ loss function \cite{DAT}. Specifically, the Adam optimizer is configured with parameters $\beta_1$ = 0.9 and $\beta_2$ = 0.99. The initial learning rate is established at $2\times10^{-4}$, with a scheduled halving at 250K, 400K, 450K, and 475K iterations. The data is augmented by random rotations of 90\degree, 180\degree, and 270\degree, as well as horizontal flipping. Implementation-wise, the model is implemented by PyTorch \cite{pytorch} and trained on a single RTX4090 GPU. 

\begin{table}[t]
\caption{This experiment employs Set5 \cite{Set5}, Set14 \cite{Set14}, B100 \cite{B100}, Urban100 \cite{Urban100}, and Manga109 \cite{Manga109} as benchmarks to compare the performance on $\times2$, $\times3$ and $\times4$ upscaling tasks of each model, thereby substantiating the effectiveness of MPSI. The results of SwinIR, ELAN, MambaIR, and DAT are the lightweight version of the model results. The evaluation criteria include PSNR and SSIM. The highest and second highest results are denoted in \textcolor{red}{red} and \textcolor{blue}{blue} fonts, respectively, for clarity and emphasis.}
\centering
\scriptsize
  \renewcommand\arraystretch{1.1} 
\begin{tabular}{|l|c|cc|cc|cc|cc|cc|}
\hline
        \rowcolor{gray!10} &  & \multicolumn{2}{c|}{Set5} &  \multicolumn{2}{c|}{Set14} &  \multicolumn{2}{c|}{B100} &  \multicolumn{2}{c|}{Urban100} & \multicolumn{2}{c|}{Manga109}  \\ 
        \rowcolor{gray!10}\multirow{-2}{*}{Method}& \multirow{-2}{*}{Scale}& PSNR & SSIM & PSNR & SSIM & PSNR & SSIM & PSNR & SSIM & PSNR & SSIM \\ \hline
        EDSR \cite{EDSR} & $\times$2 & 38.11 & 0.9602 & 33.92 & 0.9195 & 32.32 & 0.9013 & \textcolor{blue}{32.93} & 0.9351 & 39.10 & 0.9773 \\ 
        IMDN \cite{IMDN} & $\times$2 & 38.00 & 0.9605 & 33.63 & 0.9177 & 32.19 & 0.8996 & 32.17 & 0.9283 & 38.88 & 0.9774 \\ 
        LAPAR-A \cite{LAPAR} & $\times$2 & 38.01 & 0.9605 & 33.62 & 0.9183 & 32.19 & 0.8999 & 32.10 & 0.9283 & 38.67 & 0.9772 \\ 
        RDN \cite{RDN} & $\times$2 & 38.24 & 0.9614 & \textcolor{blue}{34.01} & 0.9212 & \textcolor{blue}{32.34} & 0.9017 & 32.89 & \textcolor{blue}{0.9353} & 39.18 & 0.9780 \\
        ESRT \cite{ESRT} & $\times$2 & 38.03 & 0.9600 & 33.75 & 0.9184 & 32.25 & 0.9001 & 32.58 & 0.9318 & 39.12 & 0.9774 \\
        SwinIR \cite{swinir} & $\times$2 & 38.14 & 0.9611 & 33.86 & 0.9206 & 32.31 & 0.9012 & 32.76 & 0.9340 & 39.12 & 0.9783 \\
        MambaIR & $\times$2 & 38.13 & 0.9610 & 33.95 & 0.9208 & 32.31 & 0.9013 & 32.85 & 0.9349 & 39.20 & 0.9782 \\
        ELAN \cite{ELAN} & $\times$2 & 38.17 & 0.9611 & 33.94 & 0.9207 & 32.30 & 0.9012 & 32.76 & 0.9340 & 39.11 & 0.9782 \\
        DAT \cite{DAT} & $\times$2 & \textcolor{blue}{38.24} & \textcolor{blue}{0.9614} & 34.01 & \textcolor{blue}{0.9214} & 32.34 & \textcolor{blue}{0.9019} & 32.89 & 0.9346 & \textcolor{red}{39.49} & \textcolor{red}{0.9788} \\ 
        \textbf{MPSI (ours)} & $\times$2 & \textcolor{red}{38.26} & \textcolor{red}{0.9614} & \textcolor{red}{34.02} & \textcolor{red}{0.9214} & \textcolor{red}{32.38} & \textcolor{red}{0.9022} & \textcolor{red}{33.03} & \textcolor{red}{0.9360} & \textcolor{blue}{39.47} & \textcolor{blue}{0.9787} \\ \hline
        
        EDSR \cite{EDSR} & $\times$3 & 34.65 & 0.9280 & 30.52 & 0.8462 & 29.25 & 0.8093 & 28.80 & 0.8653 & 34.17 & 0.9476 \\ 
        IMDN \cite{IMDN} & $\times$3 & 34.36 & 0.9270 & 30.32 & 0.8417 & 29.09 & 0.8046 & 28.17 & 0.8519 & 33.61 & 0.9445 \\ 
        LAPAR-A \cite{LAPAR} & $\times$3 & 34.36 & 0.9267 & 30.34 & 0.8421 & 29.11 & 0.8054 & 28.15 & 0.8523 & 33.51 & 0.9441 \\ 
        RDN \cite{RDN} & $\times$3 & 34.71 & 0.9296 & 30.57 & 0.8468 & 29.26 & 0.8093 & 28.80 & 0.8653 & 34.13 & 0.9484 \\
        ESRT \cite{ESRT} & $\times$3 & 34.42 & 0.9268 & 30.43 & 0.8433 & 29.15 & 0.8063 & 28.46 & 0.8574 & 33.95 & 0.9455 \\
        SwinIR \cite{swinir} & $\times$3 & 34.62 & 0.9289 & 30.54 & 0.8463 & 29.20 & 0.8082 & 28.66 & 0.8624 & 33.98 & 0.9478 \\
        MambaIR & $\times$3 & 34.63 & 0.9288 & 30.54 & 0.8459 & 29.23 & 0.8084 & 28.70 & 0.8631 & 34.12 & 0.9479 \\
        ELAN \cite{ELAN} & $\times$3 & 34.61 & 0.9288 & 30.55 & 0.8463 & 29.21 & 0.8081 & 28.69 & 0.8624 & 34.00 & 0.9478 \\
        DAT \cite{DAT} & $\times$3 & \textcolor{blue}{34.76} & \textcolor{blue}{0.9299} & \textcolor{blue}{30.63} & \textcolor{blue}{0.8474} & \textcolor{blue}{29.29} & \textcolor{blue}{0.8103} & \textcolor{blue}{28.89} & \textcolor{blue}{0.8666} & \textcolor{blue}{34.55} & \textcolor{blue}{0.9501} \\ 
        \textbf{MPSI (ours)} & $\times$3 & \textcolor{red}{34.79} & \textcolor{red}{0.9301} & \textcolor{red}{30.71} & \textcolor{red}{0.8487} & \textcolor{red}{29.32} & \textcolor{red}{0.8113} & \textcolor{red}{29.03} & \textcolor{red}{0.8698} & \textcolor{red}{34.64} & \textcolor{red}{0.9505} \\ \hline
        
        EDSR \cite{EDSR} & $\times$4 & 32.46 & 0.8968 & 28.80 & 0.7876 & 27.71 & 0.7420 & \textcolor{blue}{26.64} & \textcolor{blue}{0.8033} & 31.02 & 0.9148 \\ 
        IMDN \cite{IMDN} & $\times$4 & 32.21 & 0.8948 & 28.58 & 0.7811 & 27.56 & 0.7353 & 26.04 & 0.7838 & 30.45 & 0.9075 \\ 
        LAPAR-A \cite{LAPAR} & $\times$4 & 32.15 & 0.8944 & 28.61 & 0.7818 & 27.61 & 0.7366 & 26.14 & 0.7871 & 30.42 & 0.9074 \\ 
        RDN \cite{RDN} & $\times$4 & 32.47 & 0.8990 & 28.81 & 0.7871 & 27.72 & 0.7419 & 26.61 & 0.8028 & 31.00 & 0.9151 \\
        ESRT \cite{ESRT} & $\times$4 & 32.19 & 0.8947 & 28.69 & 0.7833 & 27.69 & 0.7379 & 26.39 & 0.7962 & 30.75 & 0.9100 \\
        SwinIR \cite{swinir} & $\times$4 & 32.44 & 0.8976 & 28.77 & 0.7858 & 27.69 & 0.7406 & 26.47 & 0.7980 & 30.92 & 0.9151 \\
        MambaIR & $\times$4 & 32.42 & 0.8977 & 28.74 & 0.7847 & 27.68 & 0.7400 & 26.52 & 0.7983 & 30.94 & 0.9135 \\
        ELAN \cite{ELAN} & $\times$4 & 32.43 & 0.8975 & 28.78 & 0.7858 & 27.69 & 0.7406 & 26.54 & 0.7982 & 30.92 & 0.9150 \\ 
        DAT \cite{DAT} & $\times$4 & \textcolor{blue}{32.57} & \textcolor{blue}{0.8991} & \textcolor{blue}{28.87} & \textcolor{blue}{0.7879} & \textcolor{blue}{27.74} & \textcolor{blue}{0.7428} & 26.64 & 0.8033 & \textcolor{blue}{31.37} & \textcolor{blue}{0.9178} \\ 
        \textbf{MPSI (ours)} & $\times$4 & \textcolor{red}{32.58} & \textcolor{red}{0.8991} & \textcolor{red}{28.89} & \textcolor{red}{0.7887} & \textcolor{red}{27.77} & \textcolor{red}{0.7435} & \textcolor{red}{26.71} & \textcolor{red}{0.8057} & \textcolor{red}{31.44} & \textcolor{red}{0.9181} \\ \hline
\end{tabular}
\label{table:1}
\end{table}

\subsection{Quality performance}
\subsubsection{Metrics comparison.} 
To validate the performance of MPSI, we selected several classic SR models for comparative analysis, including CARN \cite{CARN}, EDSR \cite{EDSR}, IMDN \cite{IMDN}, LAPAR-A \cite{LAPAR}, RDN \cite{RDN}, ESRT \cite{ESRT}, SwinIR \cite{swinir}, ELAN \cite{ELAN}, and DAT \cite{DAT}. Constrained by experimental conditions, a subset of the Transformer models chosen for comparison encompassed lightweight architectures, specifically SwinIR, ElAN, and DAT. Based on the results presented in Table \ref{table:1},  in the scale of $\times$2, although MPSI consistently exhibited strong performance across most validation sets, it did not exhibit an unequivocal lead over the second-best performing model. Notably, in the Manga109 dataset, MPSI's PSNR and SSIM results slightly trailed behind DAT. However, in both the $\times$3 and $\times$4 upscaling tasks, MPSI consistently achieved the highest PSNR and SSIM scores across all datasets. Particularly noteworthy is MPSI's outstanding image reconstruction performance in the Urban100 dataset. The PSNR and SSIM results of MPSI achieved on Urban100 are high in the scale of $\times$3 and $\times$4. In the scale of $\times$4, MPSI outperformed the second-best result by 0.07dB in PSNR and 0.0024 in SSIM. These outcomes underscore the superior performance of MPSI in super-resolution tasks, particularly evident in scenarios with higher upscaling factors. This can be attributed to MPSI's capability to establish comprehensive global pixel-wise sequential interaction, thereby yielding enhanced results.

\begin{figure*}[t]
\scriptsize
\centering
\scalebox{1}{
\begin{tabular}{ccc}
\hspace{-0.1cm}
\begin{adjustbox}{valign=t}
\begin{tabular}{c}
\includegraphics[width=0.318\textwidth]{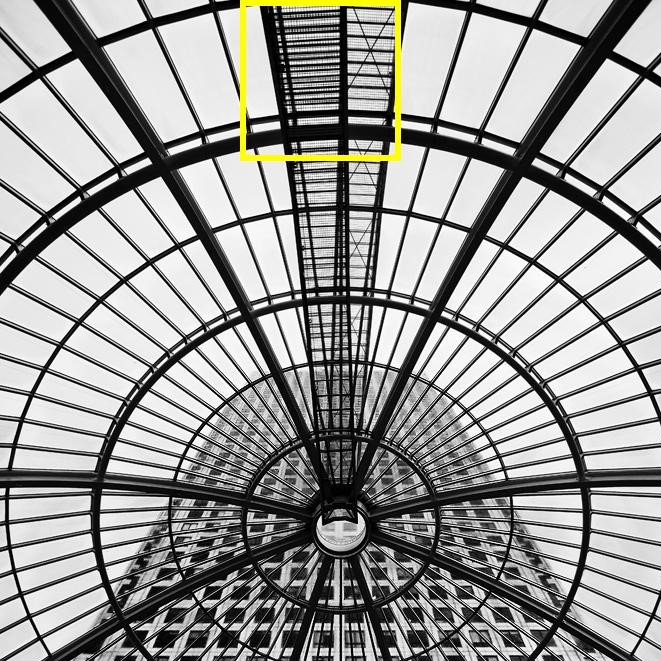}
\\
Urban100: img072
\end{tabular}
\end{adjustbox}
\hspace{-0.2cm}
\begin{adjustbox}{valign=t}
\begin{tabular}{ccccc}
\includegraphics[width=0.145\textwidth]{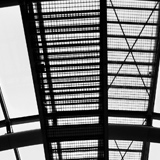} \hspace{-1mm} &
\includegraphics[width=0.145\textwidth]{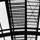} \hspace{-1mm} &
\includegraphics[width=0.145\textwidth]{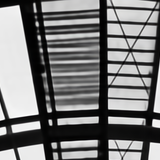} \hspace{-1mm} &
\includegraphics[width=0.145\textwidth]{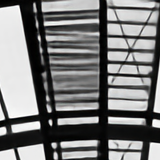} \hspace{-1mm} 
\\
HR \hspace{-1mm} &
Bicubic ($\times$4) \hspace{-1mm} &
EDSR~ \hspace{-1mm} &
IMDN~ \hspace{-1mm} 
\\
\includegraphics[width=0.145\textwidth]{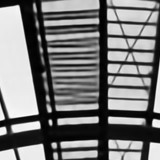} \hspace{-1mm} &
\includegraphics[width=0.145\textwidth]{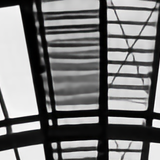} \hspace{-1mm} &
\includegraphics[width=0.145\textwidth]{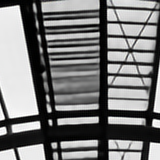} \hspace{-1mm} &
\includegraphics[width=0.145\textwidth]{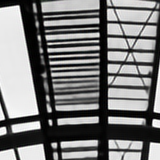} \hspace{-1mm}  
\\ 
ESRT~ \hspace{-1mm} &
SwinIR~  \hspace{-1mm} &
DAT~  \hspace{-1mm} &
MPSI \hspace{-1mm}
\\
\end{tabular}
\end{adjustbox}
\\
\hspace{-0.1cm}
\begin{adjustbox}{valign=t}
\begin{tabular}{c}
\includegraphics[width=0.318\textwidth]{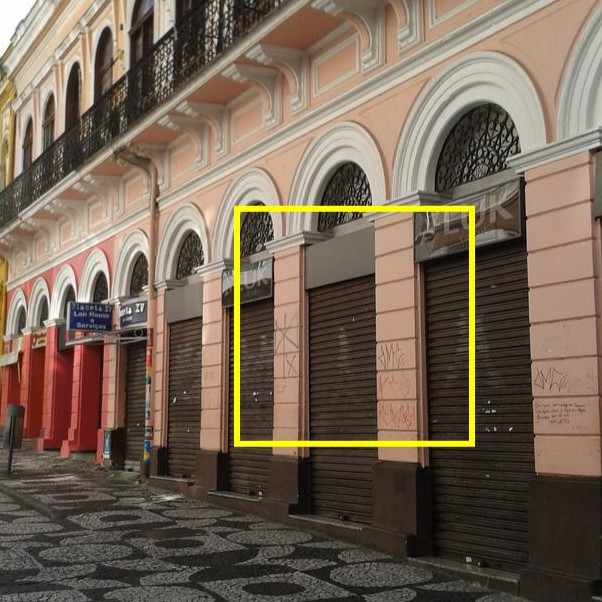}
\\
Urban100: img089
\end{tabular}
\end{adjustbox}
\hspace{-0.2cm}
\begin{adjustbox}{valign=t}
\begin{tabular}{ccccc}
\includegraphics[width=0.145\textwidth]{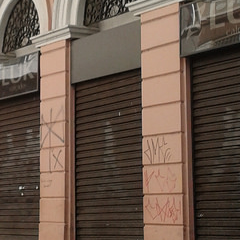} \hspace{-1mm} &
\includegraphics[width=0.145\textwidth]{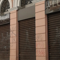} \hspace{-1mm} &
\includegraphics[width=0.145\textwidth]{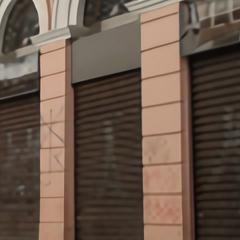} \hspace{-1mm} &
\includegraphics[width=0.145\textwidth]{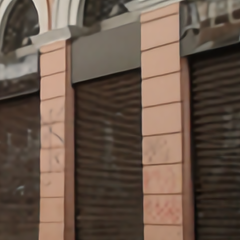} \hspace{-1mm} 
\\
HR \hspace{-1mm} &
Bicubic ($\times$4) \hspace{-1mm} &
EDSR~ \hspace{-1mm} &
IMDN~ \hspace{-1mm} 
\\
\includegraphics[width=0.145\textwidth]{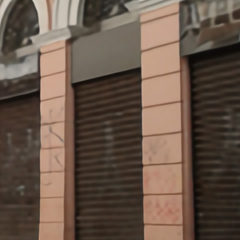} \hspace{-1mm} &
\includegraphics[width=0.145\textwidth]{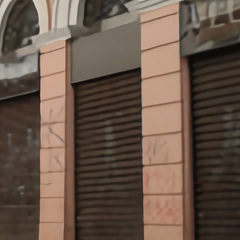} \hspace{-1mm} &
\includegraphics[width=0.145\textwidth]{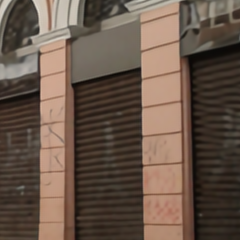} \hspace{-1mm} &
\includegraphics[width=0.145\textwidth]{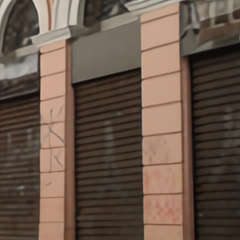} \hspace{-1mm}  
\\ 
ESRT~ \hspace{-1mm} &
SwinIR~  \hspace{-1mm} &
DAT~  \hspace{-1mm} &
MPSI \hspace{-1mm}
\\
\end{tabular}
\end{adjustbox}
\\
\hspace{-0.1cm}
\begin{adjustbox}{valign=t}
\begin{tabular}{c}
\includegraphics[width=0.318\textwidth]{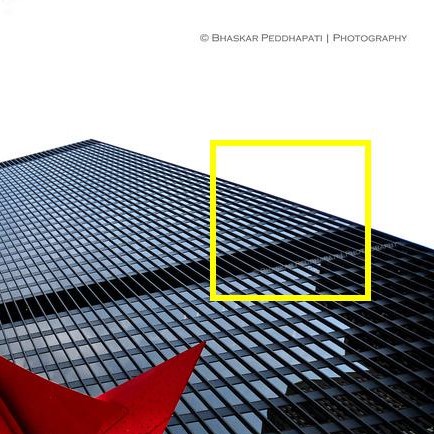}
\\
Urban100: img062
\end{tabular}
\end{adjustbox}
\hspace{-0.2cm}
\begin{adjustbox}{valign=t}
\begin{tabular}{ccccc}
\includegraphics[width=0.145\textwidth]{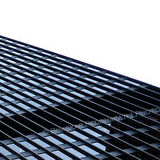} \hspace{-1mm} &
\includegraphics[width=0.145\textwidth]{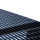} \hspace{-1mm} &
\includegraphics[width=0.145\textwidth]{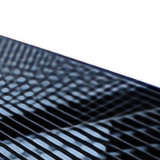} \hspace{-1mm} &
\includegraphics[width=0.145\textwidth]{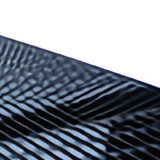} \hspace{-1mm} 
\\
HR \hspace{-1mm} &
Bicubic ($\times$4) \hspace{-1mm} &
EDSR~ \hspace{-1mm} &
IMDN~ \hspace{-1mm} 
\\
\includegraphics[width=0.145\textwidth]{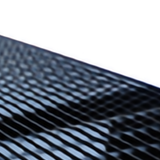} \hspace{-1mm} &
\includegraphics[width=0.145\textwidth]{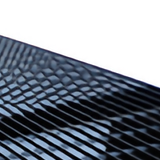} \hspace{-1mm} &
\includegraphics[width=0.145\textwidth]{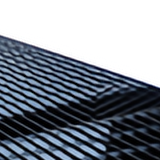} \hspace{-1mm} &
\includegraphics[width=0.145\textwidth]{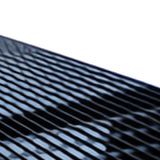} \hspace{-1mm}  
\\ 
ESRT~ \hspace{-1mm} &
SwinIR~  \hspace{-1mm} &
DAT~  \hspace{-1mm} &
MPSI \hspace{-1mm}
\\
\end{tabular}
\end{adjustbox}

\end{tabular} }
\vspace{-3mm}
\caption{\small{Visual comparison for single image SR ($\times$4 upscaling). We chose several results to conduct a comparative analysis of the restoration of details by each model. The results of SwinIR, ELAN, and DAT are the lightweight version of the model.}}
\label{fig:visual_com}
\vspace{-6mm}
\end{figure*}

\subsubsection{Visual comparison.} To assess the visual effect of image reconstruction outcomes of MPSI, we conducted a detailed comparison with the reconstruction results of several classic SR models, namely EDSR \cite{EDSR}, IMDN \cite{IMDN}, ESRT \cite{ESRT}, SwinIR \cite{swinir}, and DAT \cite{DAT}. Notably, the reconstruction results of SwinIR and DAT were obtained from their lightweight model versions. The comparison results are shown in Figure \ref{fig:visual_com}. Notably, the HR versions of each image display intricate details, yet post-bicubic processing, many of these details have become blurred. In the restoration process of image img072, while other models successfully reconstructed a majority of the local structures, certain more compact elements remained undistinguished. Conversely, the image processed by MPSI not only effectively separated the dense crossbars but also exhibited clear edges and accurate shapes. In img089, the door of the building exhibits a highly dense texture, and its dark color leads to a diminished contrast with the surrounding colors. The outcomes of other models exhibit textures that are either unclear or inaccurately arranged across different locations on the doors. Notably, MPSI's restored textures consistently demonstrated correct and clear textures across all the doors. In img062, the complexity of the texture structure in the original image is evident, with two distinct textures present in the same area, exhibiting variations in distance. Observing the results, it is evident that most models encounter challenges in accurately reconstructing the texture structure of the upper portion. DAT achieves a more precise recovery of the upper part's texture, however, there remain mistakes in the texture direction of the middle portion. In contrast, MPSI consistently restores almost all textures with higher precision while maintaining their correct orientation. The effectiveness of MPSI in this regard can be attributed to its capability to establish long-range dependence on the pixel information. This enables MPSI to enhance the global understanding of the image and facilitate detailed restoration through comprehensive feature utilization across other parts of the image.

\subsection{Ablation studies}
\begin{figure}[t]
      \centering
      \scriptsize
      \renewcommand\arraystretch{0.4} 
    \setlength{\tabcolsep}{0.4pt}
    \begin{tabular}{ccccccc}	
       \includegraphics[width=0.165\textwidth]{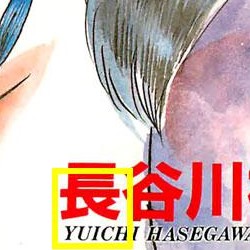} &
        \includegraphics[width=0.165\textwidth]{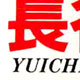} &
        \includegraphics[width=0.165\textwidth]{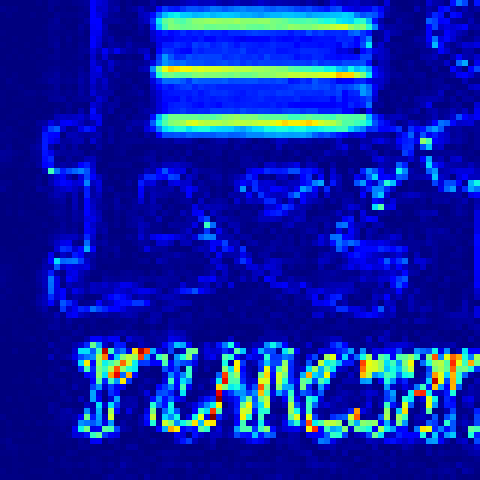} &
        \includegraphics[width=0.165\textwidth]{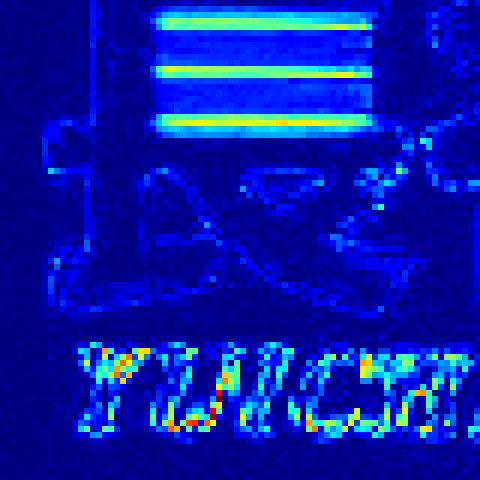} &
        \includegraphics[width=0.165\textwidth]{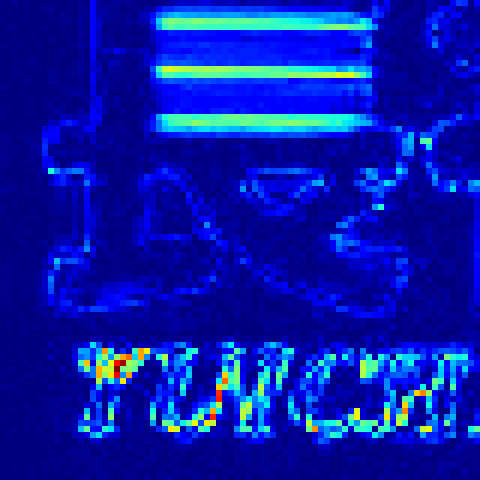} &
        \includegraphics[width=0.165\textwidth]{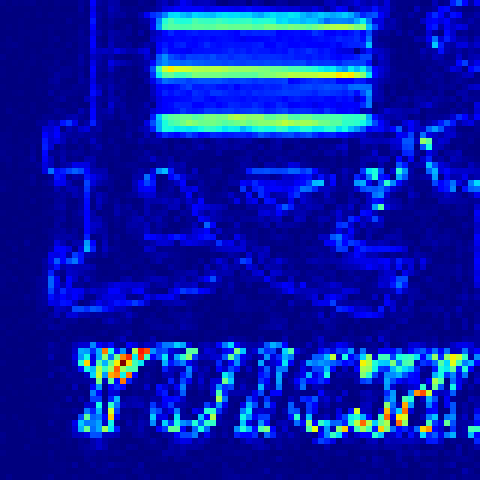} &
        \\
        \includegraphics[width=0.165\textwidth]{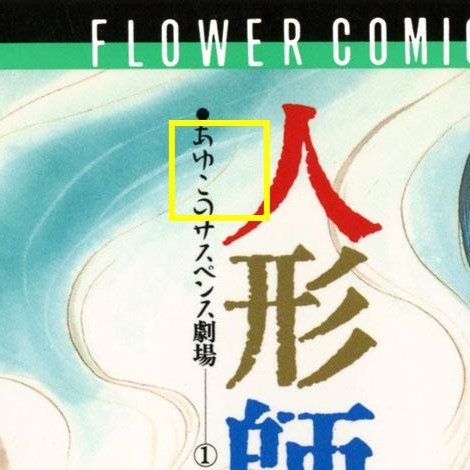} &
        \includegraphics[width=0.165\textwidth]{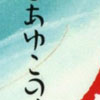} &
        \includegraphics[width=0.165\textwidth]{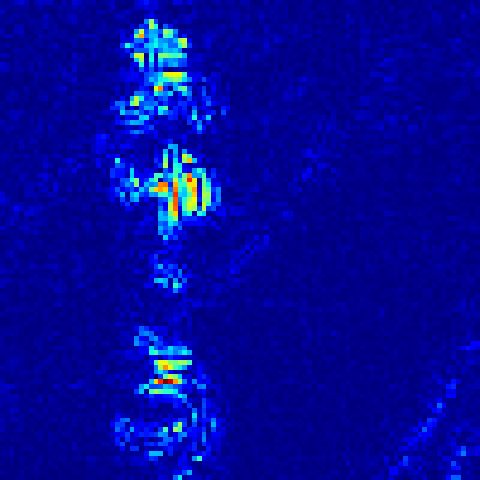} &
        \includegraphics[width=0.165\textwidth]{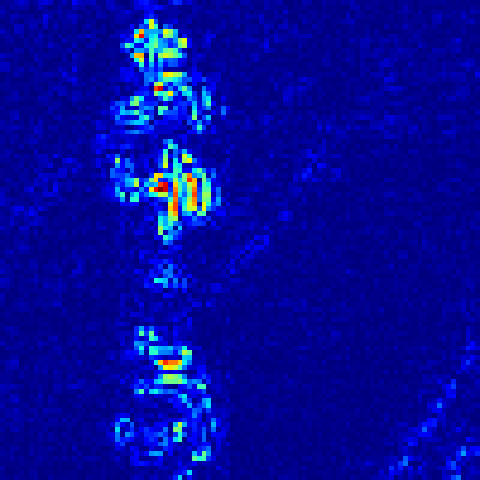}&
        \includegraphics[width=0.165\textwidth]{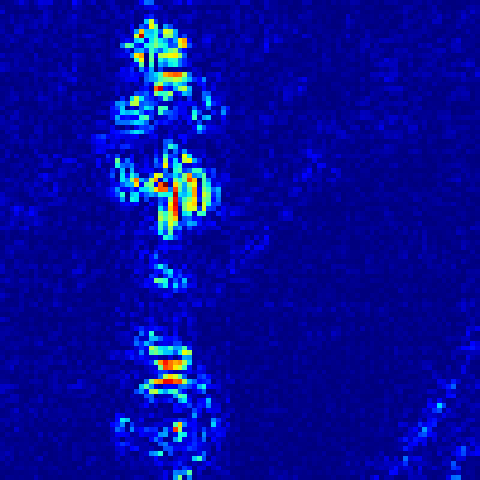}&
        \includegraphics[width=0.165\textwidth]{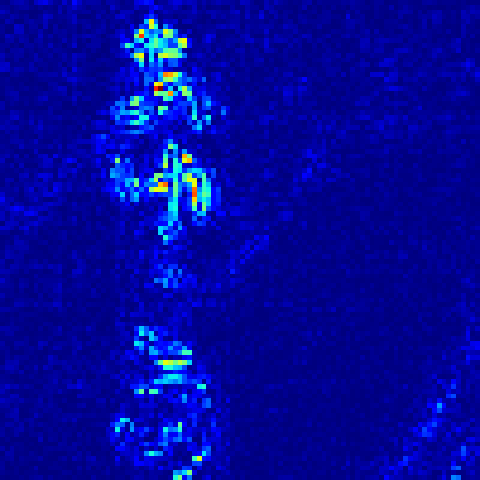}&
        \\
        HR  &
        Zoom-in&
        Baseline   &
        With CMB &
        With MCRM  &
        MPSI (ours) &
    \end{tabular}
    \vspace{-2mm}
    \caption{\small Difference maps of ablation studies on the effect of CMB and MCRM.}
    \label{fig:ablation1}
\end{figure}

\begin{figure}[t]
      \centering
      \scriptsize
      \renewcommand\arraystretch{0.4} 
    \setlength{\tabcolsep}{0.4pt}
    \begin{tabular}{cccccc}	
       \includegraphics[width=0.165\textwidth]{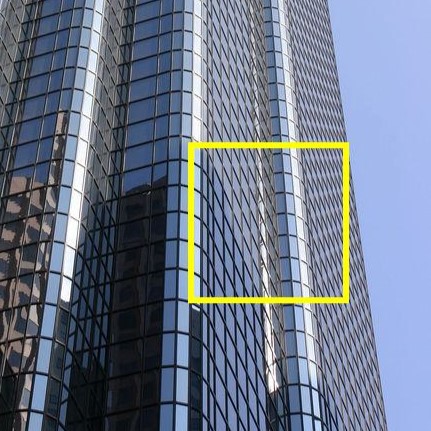} &
        \includegraphics[width=0.165\textwidth]{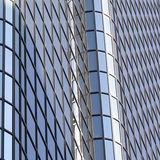} &
        \includegraphics[width=0.165\textwidth]{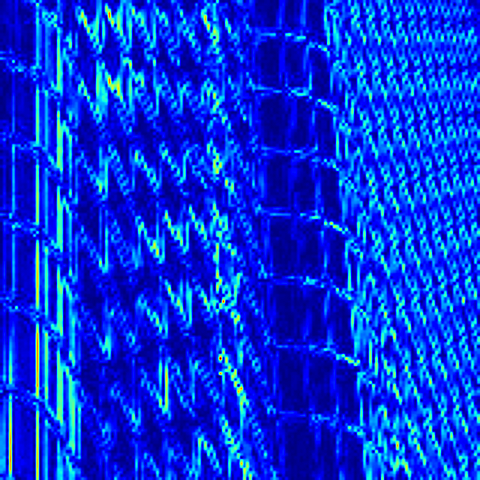} &
        \includegraphics[width=0.165\textwidth]{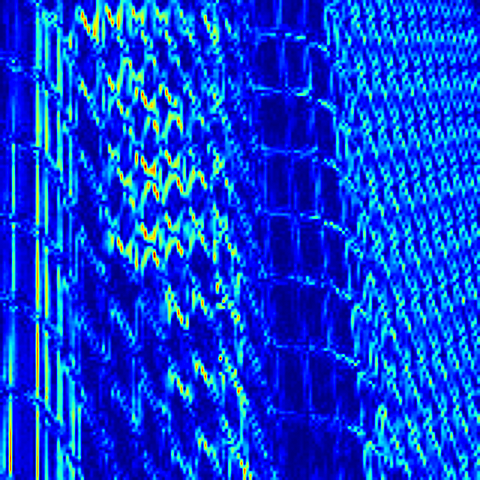} &
        \includegraphics[width=0.165\textwidth]{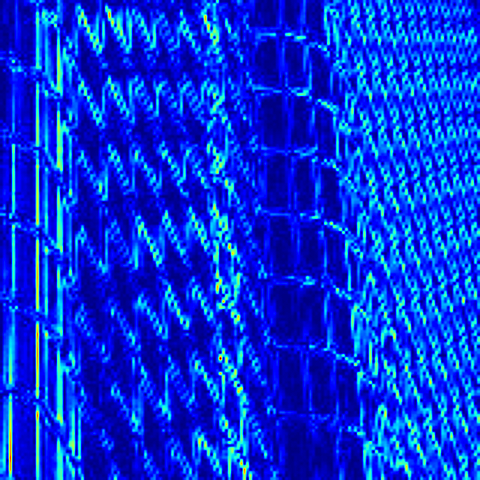} &
        \\
        \includegraphics[width=0.165\textwidth]{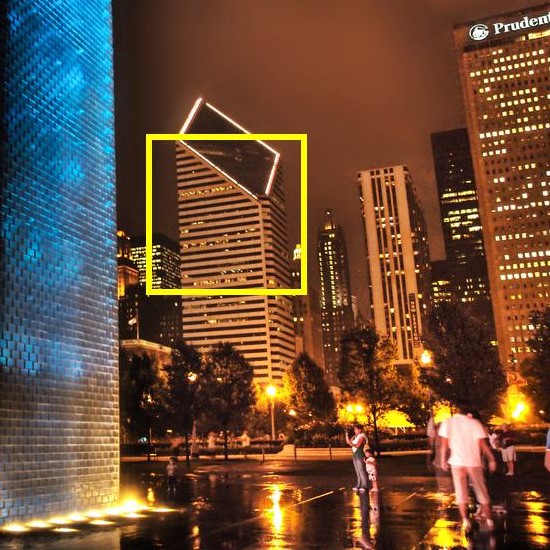} &
        \includegraphics[width=0.165\textwidth]{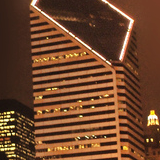} &
        \includegraphics[width=0.165\textwidth]{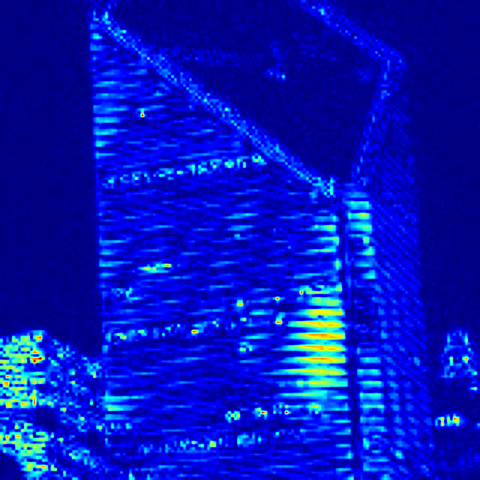} &
        \includegraphics[width=0.165\textwidth]{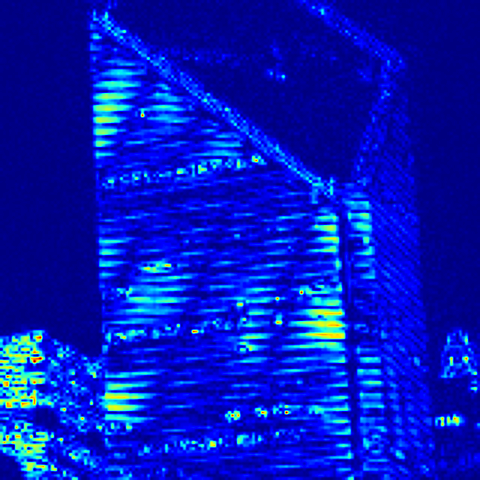} &
        \includegraphics[width=0.165\textwidth]{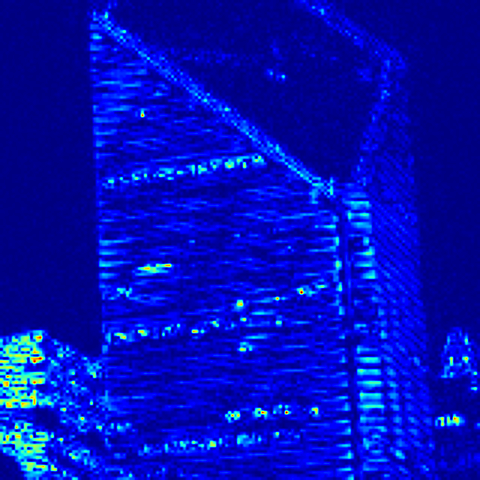} &
        \\
        HR  &
        Zoom-in&
        DDBM $\rightarrow$ CA &
        MRP $\rightarrow$ None &
        MPSI (ours) &
    \end{tabular}
    \vspace{-2mm}
    \caption{\small Difference map of ablation studies on DDBM and Mamba recursive process.}
    \label{fig:ablation2}
\end{figure}

All models are trained on the DIV2K \cite{DIV2K} and Flickr2K \cite{Flickr2K} datasets for 250K iterations, with an initial learning rate of $1\times10^{-4}$ and halving scheduled at 125K, 200K, 225K, and 237.5K iterations. The efficacy of the models is demonstrated through  PSNR and SSIM results evaluated on the Urban100 \cite{Urban100} and Manga109 \cite{Manga109} datasets, as delineated in Table \ref{table:2} and \ref{table:3}. Additionally, to underscore the model's proficiency in image restoration tasks, difference maps are meticulously crafted for selected image generation outcomes, as depicted in the accompanying Figure \ref{fig:ablation1} and \ref{fig:ablation2}. In these visual representations, regions exhibiting deeper blue hues indicate a higher similarity to the original image, whereas regions displaying darker red hues indicate more pronounced deviations from the original image.

Initially, we focus on evaluating the CMB and MCRM. The model, which excludes MCRM and replaces CMB with STB, is designated as the Baseline. For the CMB module, we replace it with STB to observe performance without CMB. Regarding MCRM, we systematically remove it from the model to assess its impact. We then investigate the internal architectures of CMB and MCRM. We substitute the DDBM within CMB with channel self-attention (DDBM $\rightarrow$ CA) to analyze its effect. To evaluate the impact of the Mamba recursive process (MRP) on MCRM performance, we retain only the features of the last layer for pooling to adjust the feature channels (MRP $\rightarrow$ None). The results are shown in Table \ref{table:2} and \ref{table:3} and Figure \ref{fig:ablation1} and \ref{fig:ablation2}.

\begin{table}[t]
\caption{\small Ablation studies on CMB and MCRM}
\vspace{-2mm}
\centering
\begin{tabular}{>{\centering\arraybackslash}b{1.5cm}>{\centering\arraybackslash}b{1.5cm}>{\centering\arraybackslash}b{1.5cm}>{\centering\arraybackslash}b{1.5cm}>{\centering\arraybackslash}b{1.5cm}>{\centering\arraybackslash}b{1.5cm}>{\centering\arraybackslash}b{1.5cm}}
        \toprule
        \rowcolor{gray!10} & & & \multicolumn{2}{c}{Urban100} & \multicolumn{2}{c}{Manga109}\\
        \rowcolor{gray!10}\multirow{-2}{*}{Base} & \multirow{-2}{*}{MCRM} & \multirow{-2}{*}{CMB} & PSNR & SSIM & PSNR & SSIM \\
        \midrule
        \checkmark & & & 26.00 & 0.7851 & 30.45 & 0.9070\\
        \checkmark & & \checkmark & 26.16 & 0.7897 & 30.66 & 0.9092\\
        \checkmark & \checkmark & &  26.01 & 0.7853 & 30.46 & 0.9072\\
        \checkmark & \checkmark & \checkmark& 26.17 & 0.7903 & 30.68 & 0.9095\\
        \bottomrule
\end{tabular}
\label{table:2}
\end{table}

\begin{table}[t]
\caption{\small Ablation studies DDBM and Mamba recursive process.}
\vspace{-2mm}
\centering
\begin{tabular}{>{\centering\arraybackslash}b{1.5cm}>{\centering\arraybackslash}b{3cm}>{\centering\arraybackslash}b{1.5cm}>{\centering\arraybackslash}b{1.5cm}>{\centering\arraybackslash}b{1.5cm}>{\centering\arraybackslash}b{1.5cm}}
        \toprule
        \rowcolor{gray!10} & & \multicolumn{2}{c}{Urban100} & \multicolumn{2}{c}{Manga109}\\
        \rowcolor{gray!10}\multirow{-2}{*}{Block} & \multirow{-2}{*}{Method} & PSNR & SSIM & PSNR & SSIM \\
        \midrule
         & DDBM $\rightarrow$ CA & 25.97 & 0.7837 & 30.42 & 0.9067\\
        \multirow{-2}{*}{CMB} & CA $\rightarrow$ DDBM & 26.17 & 0.7903 & 30.68 & 0.9091\\ \midrule
         & MRP $\rightarrow$ None & 26.14 & 0.7902 & 30.66 & 0.9089\\
        \multirow{-2}{*}{MCRM }& None $\rightarrow$ MRP & 26.17 & 0.7903 & 30.68 & 0.9091\\
        \bottomrule
\end{tabular}
\label{table:3}
\end{table}

\noindent\textbf{Ablation studies on CMB.} 
The incorporation of the CMB leads to notable improvements compared to the Baseline, as shown in the first and second rows in Table \ref{table:2}. Additionally, analysis from the figure reveals a reduction in high-difference regions within the difference maps when utilizing the CMB structure, as the fourth column shows in Figure \ref{fig:ablation1}. This shows the effectiveness of CMB in augmenting pixel-wise sequential interaction to improve reconstruction.

\noindent\textbf{Ablation studies on MCRM.} 
The incorporation of the MCRM structure leads to improvements compared to the Baseline, as shown in the first and third rows of results in Table \ref{table:2}. From the fifth column of results in Figure \ref{fig:ablation1}, it is evident that the image reconstructed with the MCRM structure exhibits notably diminished high-difference regions in the difference map. This shows that MCRM effectively enhances feature integrity, consequently improving pixel-wise sequential interaction and enhancing image reconstruction results.

\noindent\textbf{Ablation studies on CMB and MCRM.} 
As shown in the fourth row of results in Table \ref{table:2}, the model with both MCRM and CMB has the highest PSNR and SSIM results. It also has the fewest high-difference regions in the difference maps, as the sixth column shows in Figure \ref{fig:ablation1}. This indicates that the model combining CMB and MCRM structures can effectively enhance the feature extraction to improve image reconstruction results.

\noindent\textbf{Ablation studies on DDBM.} 
The first part of the data in table \ref{table:3} shows the role of the DDBM in the CMB module. When DDBM was replaced with a channel self-attention, both PSNR and SIMM dropped significantly. The comparison between the third column and the fifth column in Figure \ref{fig:ablation2} can also show that DDBM makes the difference maps have fewer high-difference areas, which shows that the super-resolution result is less different from the original image. This indicates that CMB can effectively analyze pixel information through DDBM, thereby improving pixel interaction relationships.

\noindent\textbf{Ablation studies on Mamba recursive process of MCRM.} 
The second part of the data in table \ref{table:3} shows the role of the Mamba recursive process in the MCRM module. When the model only preserves the last layer of features for adjusting the feature channels, PSNR and SSIM dropped some. The comparison of the fourth and fifth columns in figure \ref{fig:ablation2} shows that when MCRM retains the Mamba recursive process, there are fewer high-difference areas in the difference maps. The results indicate that MCRM can effectively preserve feature integrity by retaining the Mamba recursive process, leading to improved outcomes.

\section{Conclusion}
In this paper, we propose the MPSI network with the aim of improving the image reconstruction performance within single-image super-resolution tasks. We propose the CMB structure designed for efficient feature extraction and expression. Leveraging the Mamba structure, CMB facilitates the establishment of long-range dependencies among image features, enabling comprehensive capture of global image features. By amalgamating CMB with STB, SAMB can leverage the strengths of Transformer in local feature extraction and Mamba's proficiency in long-range feature dependencies, resulting in enhanced performance in image feature perception. Subsequently, we propose the utilization of Mamba to construct MCRM. This enables the amalgamation of shallower feature information into the final result during deep feature extraction, ensuring completeness of feature information. Through extensive comparative experiments, we validate the effectiveness of MPSI in super-resolution tasks, substantially improving image reconstruction results compared to previous SR models.

\clearpage  

%
%
\bibliographystyle{splncs04}
\bibliography{main}
\end{document}